\documentclass[runningheads]{llncs}
\usepackage[T1]{fontenc}
\usepackage{footnote}

\usepackage{graphicx}
\usepackage{color}
\usepackage{amsmath} 
\usepackage{booktabs} 
\usepackage{array}    
\usepackage{tabularx}
\usepackage{float}
\usepackage{multirow} 
\usepackage{adjustbox}

\usepackage{pgfplots}
\pgfplotsset{compat=1.16} 

\usepackage{lipsum} 
\usepackage{authblk}

\begin{document}
%
\title{Enhancing Financial Market Predictions: Causality-Driven Feature Selection}
%
%
\author{Wenhao Liang\and
Zhengyang Li\and
Weitong Chen \thanks{Corresponding Author. Email: Weitong.Chen@Adelaide.edu.au}}
%
\institute{{School of Computer and Mathematical Sciences \\ Adelaide University, Adelaide, Australia\\}
}
\maketitle              
\begin{abstract}
This paper introduces FinSen dataset that revolutionizes financial market analysis by integrating economic and financial news articles from 197 countries with stock market data. The dataset’s extensive coverage spans 15 years from 2007 to 2023 with temporal information, offering a rich, global perspective 160,000 records on financial market news. Our study leverages causally validated sentiment scores and LSTM models to enhance market forecast accuracy and reliability. Utilizing the FinSen dataset, we introduce an innovative Focal Calibration Loss, reducing Expected Calibration Error (ECE) to 3.34\% with the DAN 3 model. This is not only improves prediction accuracy but also aligns probabilistic forecasts closely with real outcomes, crucial for financial sector where predicted probability is paramount. Our approach demonstrates the effectiveness of combining sentiment analysis with precise calibration techniques for trustworthy financial forecasting where the cost of misinterpretation can be high.

\keywords{Financial News Dataset \and Time-Series \and Calibration }

\end{abstract}
%
%
%

\section{Introduction}
Forecasting stock market volatility  \cite{ge2022neural,poon2003forecasting,vui2013review} presents a paramount yet formidable challenge within the financial sector, attributed to the market's non-linear and inherently volatile nature. The capacity to predict market trends not only underpins investment strategies but also informs economic forecasting and risk management practices. The complexity of achieving accurate predictions is exacerbated by the market's sensitivity to temporal factors, with prices reflecting the latest events and sentiments  \cite{baker2007,solaiman2024unveiling}. This sensitivity leads to rapid and often unpredictable fluctuations, influenced by a confluence of factors including economic indicators, global events, and investor behaviors. 

Traditional models  \cite{bao2017deep,engle2012forecasting,franses1996forecasting,gong2019forecasting,marcucci2005forecasting}, which predominantly rely on historical numeric data, encounter limitations in capturing the multifaceted dynamics of market volatility. Moreover, a notable gap exists in public datasets \cite{reuters-21578_text_categorization_collection_137,malo2014good,pei2022tweetfinsent,PsychSignal,zhang2015character}, particularly in their lack of precise temporal information, complicating the pursuit of unknown factors—referred to as alpha or beta that significantly impact financial market volatility. The Efficient Market Hypothesis (EMH) \cite{fama1970efficient} is one of the key theories in behavioral finance, which set the stage for considering alternative indicators like sentiment in market analysis. This context in Table~\ref{tab1} raises the pivotal research question: Is the contribution of sentiment from news events a reliable predictor of market volatility?

\begin{table}
\caption{High-Tech Companies News and Movement (Mov.) from June 2022 to June 2023. Sentiments are represented by positive and negative.}\label{tab1}
\begin{tabular}{|l|c|c|p{7cm}|l|}
\hline
Date	&Symbol	&Mov.(\%)	& News Events & Sent.\\
\hline
2022/7/27	&	MSFT	&	+6.68	&	Microsoft Cloud strength drives fourth quarter. & Pos.\\
2022/11/30	&	MSFT	&	+6.16	&	ChatGPT has launched on November 30, 2022. & Pos.	\\
2023/1/20	&	GOOG	&	-5.72	&	Eliminating about 12000 jobs, or 6\% of workforce.	& Neg.	\\
2022/11/10	&	META	&	-10.25	& Mark Zuckerberg's Meta cuts 11000 jobs.	& Neg.\\
2022/10/28	&	AAPL	&	+7.56	&	Titan's sales and profits exceeded expectations.	& Pos.\\
\hline
\end{tabular}
\end{table}


Antweiler and Frank's \cite{antweiler2004all} seminal study analyzed the influence of internet stock message boards on market volatility using traditional statistical methods, which struggled to capture market dynamics' non-linearity. Conversely, Neural Networks (NNs) have shown superior capability in detecting the non-linear nature of stock market volatility \cite{ge2022neural,vui2013review}. Furthermore, the incorporation of sentiment analysis from unstructured financial news sources allows neural networks (NNs) to comprehend more nuanced market sentiments, enhancing their predictive accuracy in financial markets. Subsequent research, such as Li et al.'s  \cite{li2014news} examination of the effects of news sentiment on stock prices, illustrates the benefits of merging NNs with sentiment analysis to refine prediction accuracy. 
Neural networks excel at identifying patterns and correlations within datasets, yet they often overlook the temporal sensitivity inherent in stock market data. Contemporarily, deep learning techniques, including LSTMs \cite{jin2020stock,mohan2019stock} and Transformer-based models like BERT \cite{hiew2019bert}, which have been refined by incorporating sentiment analysis for superior context and nuance understanding. However, they do not inherently understand the causality behind these correlations that make NNs model applied with less interpretability. In addition, modern NNs are poorly calibrated even though they have higher accuracy \cite{guo2017calibration,niculescu2005predicting}. In critical decision-making fields such as financial investment \cite{liu2019neural}, a poor calibrated model can potentially lead to lower confidence on prediction, making an unreliable investment in real-world scenarios. We consider a truth-worthy model is much more significant than other SOTA model in the context of financial prediction.

To close the gaps above, our work publicly presents a novel financial dataset, FinSen, notable for its temporal attributes. Subsequently, utilizing the FinSen dataset, we establish a demonstrable causal relationship between market sentiment scores and volatility within the S\&P 500. Thirdly, capitalizing on the established causality, we enrich LSTM models by integrating causal-validated sentiment scores, yielding impressive results and interpretability to financial prediction model. Lastly, we propose a novel loss function, Focal Calibration Loss, to further enhance trustworthiness and dramatically improves the Expected Calibration Error (ECE) to 3.34\% upon the SOTA calibration methods, signifying a close alignment of probabilistic predictions with actual market outcomes.

\section{Related Works}
\subsection{Related Sentiment Datasets on Stock Market Prediction}

The predictive modeling of stock prices has been a focal point of many studies, leveraging a wide array of data characteristics. Among these, sentiment datasets derived from financial news have been identified as particularly valuable for capturing the market's mood and its potential impact on stock volatility. \textbf{The Financial PhraseBank} \cite{malo2014good} contains sentences from financial news categorized into positive, negative, and neutral sentiments, specifically designed for financial context. \textbf{Reuters-21578 News Dataset} \cite{reuters-21578_text_categorization_collection_137} is a foundational benchmark for text classification tasks, originating from the Reuters newswire service in the year 1987. Comprising a total of 21,578 documents, but it is no longer available online for research purposes due to copyright issues. \textbf{StockTwits Dataset} \cite{batra2018integrating,jaggi2021text,oliveira2013predictability,oliveira2014automatic} includes messages from the StockTwits platform, tagged with sentiments by users, providing a crowd-sourced sentiment analysis on stocks . \textbf{TweetFinSent} \cite{pei2022tweetfinsent} is dataset of stock sentiments on Twitter,  designed for sentiment analysis in the stock market domain, with a focus on "meme stocks.", which is distinctive due to its expert-annotated nature. \textbf{PsychSignal Dataset} \cite{PsychSignal} gathers data from multiple sources, including Twitter, StockTwits, and various of market sentiments across different social media and online platforms. \textbf{AG News} \cite{zhang2015character} is  commonly used in text classification tasks and consists of news articles from the AG's corpus of news articles on the web, categorized into four main topics: World, Sports, Business, and Science. While these datasets ignore the influences of financial indicator and assume that the stock market price can be easily predicted by stock price and the news itself. We argued that financial indicator and news are main contributors.

\begin{table}[ht]
\small
\centering

\caption{Comparison of Financial Sentiment Datasets}
\label{tab:dataset-comparison}
\resizebox{0.8\textwidth}{!}{%
\begin{tabular}{c|c|c|c|c}

\hline
\textbf{Dataset} & \textbf{Scope} & \textbf{Size} & \textbf{Annotation}  & \textbf{Temperol} \\ \hline
PhraseBank & Finance News  & 4840 & Expert & N \\ \hline
Reuters-21578 & Finance News & 21,578 & User   & 1Y, 1987 \\ \hline
StockTwits & Stock Discussions & 6.4M & User& 10Y \\ \hline
TweetFinSent & Meme Stocks   & 2,113 & Expert   & N \\ \hline
PsychSignal & Market News & Varied & Algorithm \& User   & N \\ \hline
AG News & News & 1M  & User & N \\ \hline
FinSen (Ours) & Market News \& Indicators & 160K & FinBERT & 15Y \\ \hline
\end{tabular}
}
\end{table}

\subsection{Related Stock Market Forecasting Methods}
The fusion of sentiment analysis and machine learning techniques has emerged as a promising avenue in the realm of stock market prediction. While the LSTM model with an attention mechanism \cite{li2017sentiment} has proven its superiority in prediction accuracy, it falls short in providing reliable confidence estimates. Although previous works have shown the effectiveness of various techniques on financial prediction, they all ignore model interpretability and trustworthiness, risking the failing rate of investment in practical market. The KGEB-CNN technique\cite{ding2016knowledge}, utilized data from the S\&P 500 and Reuters, achieving an accuracy of 66.93\%. Similarly, Vargas et al.\cite{vargas2017deep} used the SI-RCNN technique with data from the S\&P 500 and Reuters, yielding an accuracy of 62.03\%. Chen et al.\cite{chen2019incorporating} implemented the BiLSTM-CRF technique on data from TOPIX and Reuters, resulting in an accuracy of 65.7\%. Deng et al.\cite{deng2019knowledge} applied the KDTCN technique to DJIA and Reddit WorldNews data, achieving the highest accuracy of 71.8\%. Additionally, Jin et al.\cite{jin2020stock} used the EMD-LSTM technique with data from Apple and StockTwits, reaching an accuracy of 70.01\%. Li et al.\cite{li2020incorporating} employed the LSTM technique on data from the Hong Kong Exchange and FINET, with an accuracy of 60.10\%. Finally, Li et al.\cite{li2017sentiment} also used the LSTM technique, but with data from CSI300 and Eastmoney Forum, achieving the highest reported accuracy of 87.86\%.

\subsection{Nerual Network Model Calibration Techniques}

Initial efforts in model calibration, such as Platt Scaling \cite{platt1999probabilistic} and Isotonic Regression \cite{zadrozny2001obtaining}, primarily focused on adjusting the outputs of already trained models. These techniques, while pioneering, often fell short in directly influencing the model’s internal decision-making process, propelling the development of more flexible approaches. Later modification on loss function, Focal Loss \cite{lin2017focal} potentially leads to overconfident or underconfident predictions because it modifies the standard cross-entropy loss to focus more on hard, misclassified examples, without directly addressing how accurate the predicted probabilities are in reflecting the true likelihood of an event \cite{ghosh2022adafocal,tao2023dual}. Various network calibration techniques are measured by sophisticated metrics that gauge calibration quality, such as the Expected Calibration Error (ECE) \cite{guo2017calibration} and Brier Score \cite{brier1950verification}.

\section{Data and Methodology}
\begin{figure}[ht]
\centering
\includegraphics[width=0.8\textwidth]{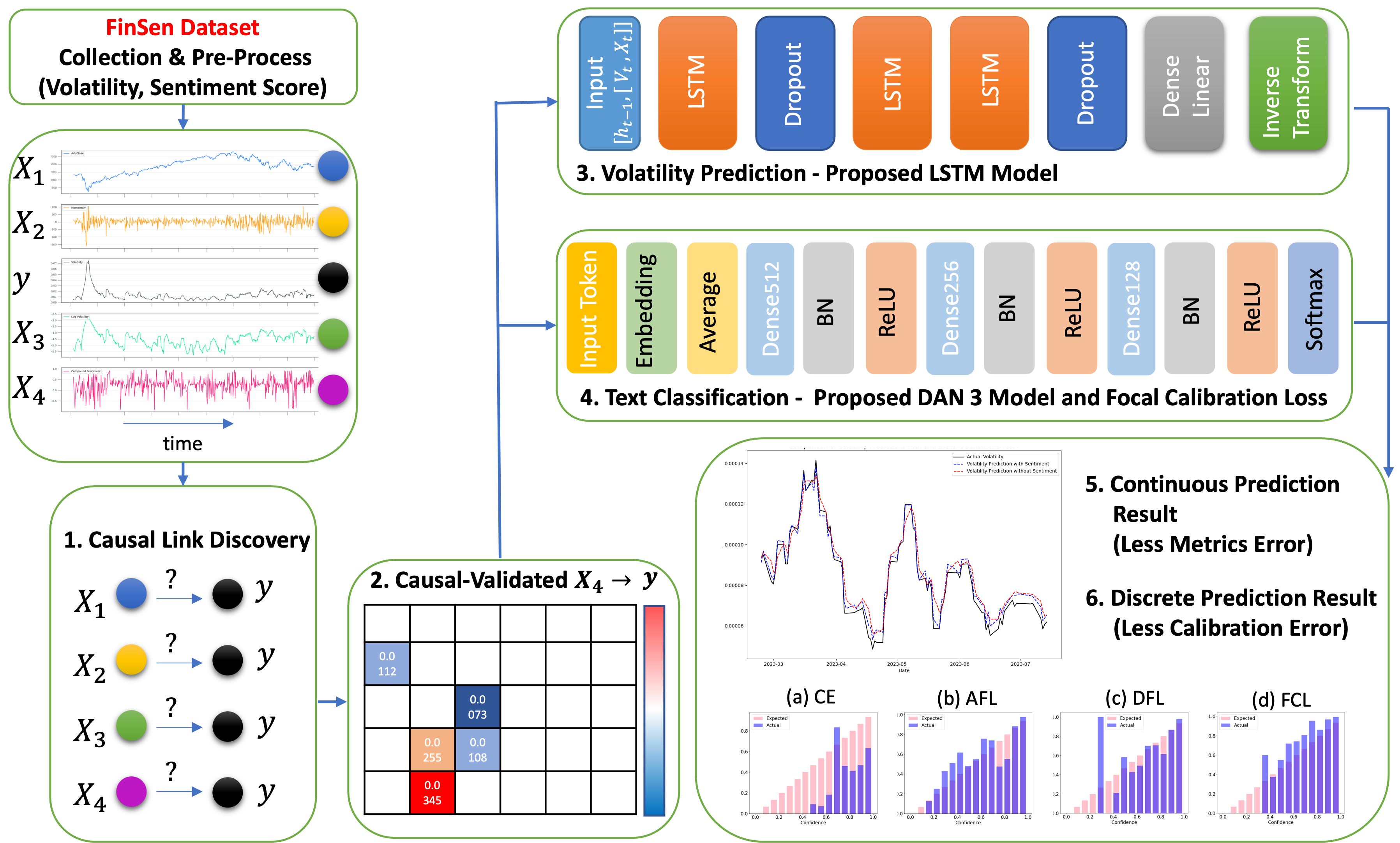}
\caption{Framework of LSTM Volatility Prediction and DAN 3 Text Classification. \(X_4\) is sentiment scores generated from FinSen dataset by FinBERT as the input of both models after causally validated with volatility \(y\).} \label{fig:framework}
\end{figure}

\begin{figure}
\centering
\includegraphics[width=0.8\textwidth]{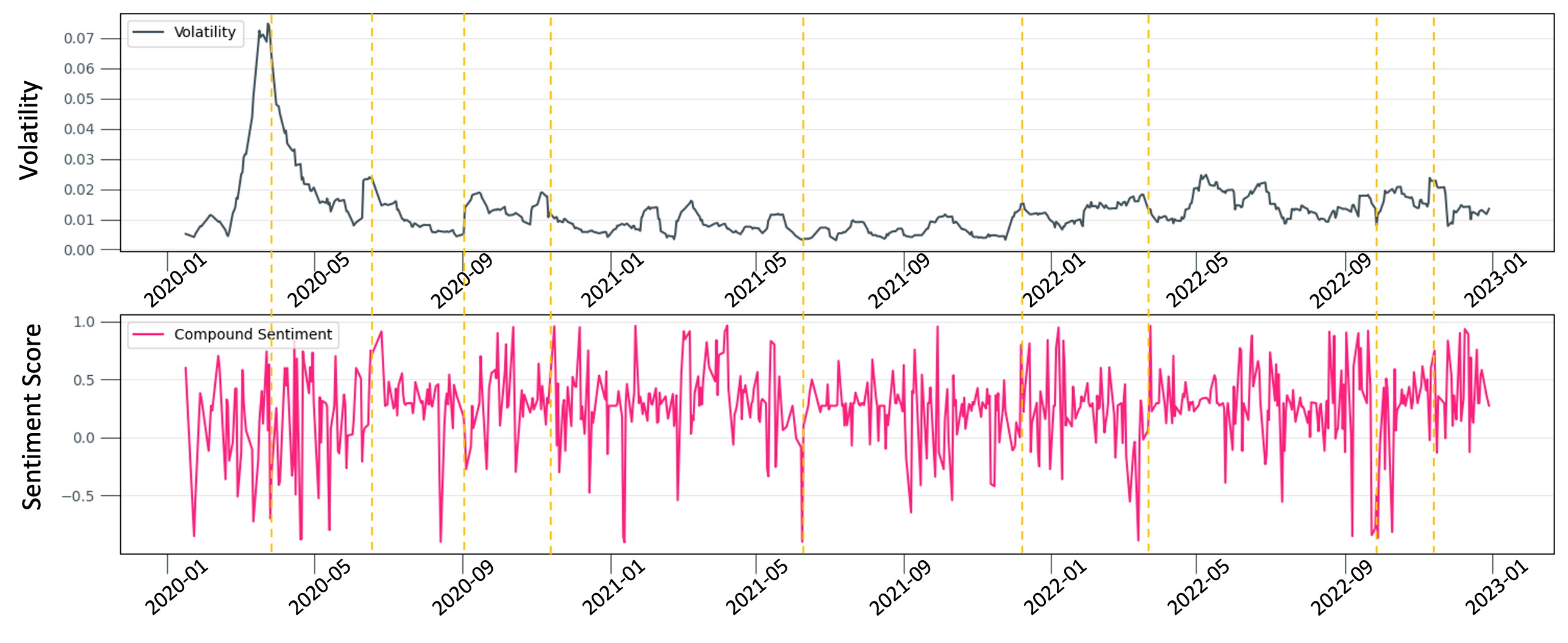}
\caption{Granger Cause Test: \(X\) is Aggregated Sentiment Scores and \(y\) is Market Volatility. Yellow dashed lines highlight prominent peaks and valleys of alignment, selected subjectively to illustrate instances of notable alignment between features.\protect\footnotemark} \label{fig:gc}
\end{figure}

The target variable is the daily percentage change in the Standard \& Poor’s 500 Index (S\&P 500), which serves as an indicator of the US stock market's health. We utilize the volatility derived from its closing price to explore the causal relationship with sentiment scores from our FinSen(US) dataset, as illustrated in Fig.~\ref{fig:gc}. Upon establishing causal evidence, we incorporate this causally validated feature into our proposed LSTM model for prediction and calibration experiments as shown in Fig.~\ref{fig:framework}.

\subsection{FinSen Data Collection and Preprocessing}

\textbf{Numerical Data} The S\&P 500 OHLCV data, including open, high, low, close, and volume, was used to extract price momentum and volatility. This data was sourced from the Yahoo Finance API \cite{yhoo} from January 2020 to June 2023. Our study's supervised target are the S\&P 500 index momentum \cite{Wilder1978}, volatility \cite{Agustini2018GeometricBrownian,Markowitz1952}, and log volatility \cite{Agustini2018GeometricBrownian} which are defined in Appendix 1. \textbf{Textual Data} Data collection was conducted using advanced web scraping techniques to collect news headlines and contents from reputable financial news website, Trading Economics, source of economic indicators and financial news  \cite{TradingEconomics}. The Trading Economics platform offers access to over 20 million indicators from 196 countries, including historical and current data on exchange rates, stocks, indexes, bonds, and commodity prices. Our FinSen dataset, comprising over 160,000 records collected over a span of 15 years, was curated to ensure a broad representation of global economic conditions with temporal information. 
Sentiment analysis is conducted 
 and annotated on the preprocessed financial news articles using FinBERT \cite{devlin2018bert}, a finetuned BERT model specifically designed for financial text. This step transforms qualitative news content into quantitative sentiment scores, facilitating the integration of sentiment data. The model and its tokenizer are accessed from the Hugging Face Transformers library  \cite{wolf2020transformers} under the identifier "ProsusAI/finbert". Table~\ref{tabfinbert}, on a few selected days, the sentiment scores we obtained from FinBERT.

\begin{enumerate}
    

    \item Sentiment Scoring of News: The encoded text is fed into FinBERT, which outputs logits representing the sentiment classification. This score, denoted as \( S_i \), ranges possibilities among -1 (negative), 0 (neutral) and +1 (positive), representing the sentiment of the \( i^{th} \) news.
    
    \item Daily Sentiment Aggregation: To obtain a daily sentiment score, we aggregated the sentiment scores of all news for each day. The daily sentiment score \(SAgg_d = \frac{1}{N_d} \sum_{i=1}^{N_d}( (-1*P_{neg})+(0*P_{neu})+(1*P_{pos}))\), for day \( d \) is calculated as the average of sentiment scores of all news made on that day, where \( N_d \) is the total number of news on day \( d \).

\end{enumerate}

\begin{table}[ht]
\centering
\caption{FinSen Dataset: The Aggregated Sentiment Scores by FinBERT}\label{tabfinbert}
\resizebox{0.7\textwidth}{!}{%
\begin{tabular}{|l|p{5.5cm}|l|p{1.7cm}|p{1cm}|}
\hline
\textbf{Date}& \textbf{News Content}& \textbf{Sent.}& \textbf{FinBERT}& $\mathbf{SAgg_d}$\\
\hline
2023/2/14	&	Coca-Cola earnings meet market expectations at 0.45 USD. & Pos.& [0.448771, 0.0296302, 0.5215982] & 0.6597\\
2023/3/10	& The unemployment rate in the US edged up to 3.6 percent in February 2023, up from a 50-year low of 3.4 percent seen in January and above market expectations of 3.4 percent. & Neg.& [0.9373965, 0.04212248, 0.02048101] & 0.2637\\
2023/4/25	 &Microsoft decreased to a 4-week low of 278.39. & Neu.& [0.01425993, 0.9658618, 0.01987824] & 0.2142\\
2023/5/03	   & Apple increased to a 35-week high of 170.47. & Neg.& [0.9287658, 0.02683192, 0.04440225] & 0.2293\\

\hline
\end{tabular}
}
\end{table}

\footnotetext{It is important to note that the stock market is influenced by various factors beyond investor sentiments, including economic indicators, geopolitical events, and market speculation\cite{baker2007}. While sentiment scores offer valuable insights, they do not solely dictate market fluctuations, and market reactions to news can be immediate or delayed. The presence of unaccounted variables can lead to instances where alignment between sentiment scores and market volatility is not perfect.}

\subsection{Proposed Methods}
\textbf{Integrating Causal-Validated Sentiment Scores into LSTM for Volatility Prediction} To explore the potential of market sentiment as a predictive factor for S\&P 500 volatility, we conducted a Granger causality test \cite{Granger1969} and treated \( SAgg_d \) as \(X\). This statistical method determines whether one time series can forecast another, effectively testing for causality between two variables. Specifically, we assessed whether sentiment scores (\(X\)) derived from financial news could predict the S\&P 500's volatility (\(y\)) by equation (\ref{eq:gc}). If the coefficients \(\lambda_i \) are statistically significant, \(X\) Granger-causes \(y\). 
\begin{gather}
y_t = \alpha + \sum_{i=1}^k \beta_i Y_{t-i} + \sum_{i=1}^k \lambda_i X_{t-i} + \epsilon_t \label{eq:gc}\\
X_{t} = \{X | X \, \text{Granger-causes}\, y\}.\label{eq:gcxy}
\end{gather}
Given the established Granger causality between market sentiment scores (\(X\)) and the S\&P 500's volatility (\(y\)), we integrated sentiment analysis into an LSTM model to predict the volatility. The LSTM model uses the traditional input features along with sentiment scores, represented by the feature vector \([V_t,X_t]\), where \(V_t\) denotes the 'Volatility' feature and \(X_t\) represents the sentiment scores at time step $t$. The LSTM model is based on the standard LSTM structure defined as \(h_t = \text{LSTM}(h_{t-1}, [V_t, X_t])\label{eq:lstm1}\), where $h_t$ is the hidden state at time $t$, and LSTM represents the LSTM function that updates the hidden state based on the previous hidden state $h_{t-1}$ and the input features \([V_t,X_t]\).

\noindent \textbf{Embedding Causal-Validated Feature into DAN 3 for Text Classification} Our approach implemented 3 feed-forward layers of Deep Averaging Network with Batch Normalization (DAN 3) \cite{guo2017calibration}, which is structured in Fig.~\ref{fig:framework} and to process textual data by first converting token indices to dense embeddings, followed by averaging these embeddings to form a fixed-size vector representation of each text instance. This representation is then processed through a series of dense layers interleaved with batch normalization, culminating in an output layer that maps to the number of output classes.

\noindent \textbf{Focal Calibration Loss} Our Focal Calibration Loss is designed to address two main challenges for text classifier during training: class imbalance and model calibration. The intuition behind its design is rooted in enhancing model performance while ensuring the predicted probabilities are aligned with the true outcomes. The Focal Calibration Loss (Eq.~\ref{eq:fcl}) combines with two components, Focal Loss \cite{lin2017focal} and Calibration Error (Eq.~\ref{eq:ce}), incorporating a regularization parameter \(\lambda\) to balance the trade-off between focusing on hard examples and ensuring proper calibration of the model's predictions:
\begin{small}
\begin{gather}
\text{FL}(p_t) = -(1 - p_t)^\gamma \log(p_t)
,\quad \text{CalibError} = \frac{1}{N} \sum_{i=1}^{N} \sum_{c=1}^{C} |p_{i,c} - y_{i,c}| 
\label{eq:ce}
\\
\text{FocalCalibrationLoss} = \text{FL}(p_t) + \lambda \cdot \text{CalibError} 
\label{eq:fcl}
\end{gather}
\end{small}
In Focal Loss term, \(p_t\) is the model's estimated probability for the class label \(t\). \(\gamma\) is the focusing parameter that adjusts the rate at which easy examples are down-weighted. In Calibration Error term, formulated as the mean absolute error between the predicted probabilities and the one-hot encoded ground truth labels across all classes. \(N\), \(C\)  is the number of examples and classes, \(p_{i,c}\) is the predicted probability of class \(c\) for the \(i\)th sample, \(y_{i,c}\) is the one-hot encoding of the true class label for the \(i\)th example. The theoretical underpinning of using MAE for Calibration Error comes from the concept of expected calibration error (ECE) \cite{guo2017calibration}, which is a popular metric for assessing calibration by grouping predictions into bins and comparing the average predicted probability to the actual outcome frequency within each bin. While ECE provides a binned estimate of calibration, MAE offers a continuous, instance-level calibration measure.

\section{Experiments and Results}

\subsection{Establishing Causality: Sentiment Scores on Market Volatility}
\begin{small}
\begin{gather}
    \Delta \text{SP500}_t = \begin{cases} 
    a_0 + \sum_{i=1}^{k} a_i(\Delta \text{SP500})_{t-i} + \sum_{i=1}^{k} b_i(X_{t-i}) + \epsilon_t \\
    c_0 + \sum_{i=1}^{k} c_i(\Delta \text{SP500})_{t-i} + \sum_{i=1}^{k} d_i(X_{t-i}) + \epsilon_t
    \label{eq:gc1gc2}
    \end{cases}\\
    F = \frac{(SSR_R - SSR_U) / p}{SSR_U / (T - k - 1)}.
    \label{eq:gc5}
\end{gather}
\end{small}

Our analysis begins with evaluating the stationarity of the percent changes in both the volatility of S\&P 500  and the news sentiment scores. We applied Augmented Dickey-Fuller (ADF) Test \cite{dickey1979distribution} that they are stationary and suitable for Granger causality testing. To determine the causal relationship, we employ a regression model that predicts the percentage change in the S\&P 500 , denoted as \( \Delta \text{SP500} \), based on sentiment scores from financial news articles. The model is structured to include the sentiment scores lagged up to 30 days, represented as \(X_{t-i}\), where \(i\) ranges from 1 to \(k\), with \(k\) being the maximum lag of 30 days. 






The null hypothesis posits that the coefficients $b_i = 0$ for all $i$. We apply an F-test by equation(\ref{eq:gc5}) to the residuals from the restricted model (excluding sentiment scores) and the unrestricted model (including sentiment scores). If the F-value is greater than the critical value at a chosen level of significance (0.05), we reject the null hypothesis, suggesting that sentiment scores "Granger cause" the S\&P 500  volatility. The null hypothesis for each test is that the lagged values of the independent variable do not belong in the regression. Based on the results from these tests, we formulated four possibilities representing potential causal relationships between sentiment scores and \( \Delta \)SP500:



\begin{enumerate}
    \item Sentiment scores "Granger-cause" \( \Delta \)SP500 if they improve the prediction of volatility, and volatility do not improve the prediction of sentiment scores ($b_i \neq 0$ and $d_i = 0$).
    \item \( \Delta \)SP500 "Granger-causes" sentiment scores if volatility improve the prediction of sentiment scores, and sentiment scores do not improve the prediction of volatility ($b_i = 0$ and $d_i \neq 0$).
    \item A mutual relationship exists if both sentiment scores "Granger-cause" \( \Delta \)SP500, and \( \Delta \)SP500 "Granger-causes" sentiment scores ($b_i \neq 0$ and $d_i \neq 0$).
    \item Independence is indicated when no causal relationships are found ($b_i = 0$ and $d_i = 0$).
\end{enumerate}

The results of the Granger causality tests for Eq.(\ref{eq:gc1gc2}) are summarized in Fig.~\ref{fig:granger}. We tested whether \( \Delta \)SP500 predicts sentiment score, and whether sentiment score predicts \( \Delta \)SP500. Separate regressions were run for all values of \( k \) (1 to 30), where \( k \) represents the maximum lag length in the regression.  The F-statistics are significant at the 1\%, 5\%, and 10\% levels for lagged day 1, 3, 7, 14 and 30, respectively. This suggests that sentiment score does "Granger cause" \( \Delta \)SP500. However, none of the F-statistics are sufficient to reject the null hypothesis, indicating that \( \Delta \)SP500 does not "Granger cause" sentiment score. The results suggest that sentiment scores do "Granger cause" volatility. 


\subsection{Enhancing LSTM with Causal-Validated Sentiment Scores}

\begin{figure}[ht]
\centering
\includegraphics[width=0.7\textwidth]{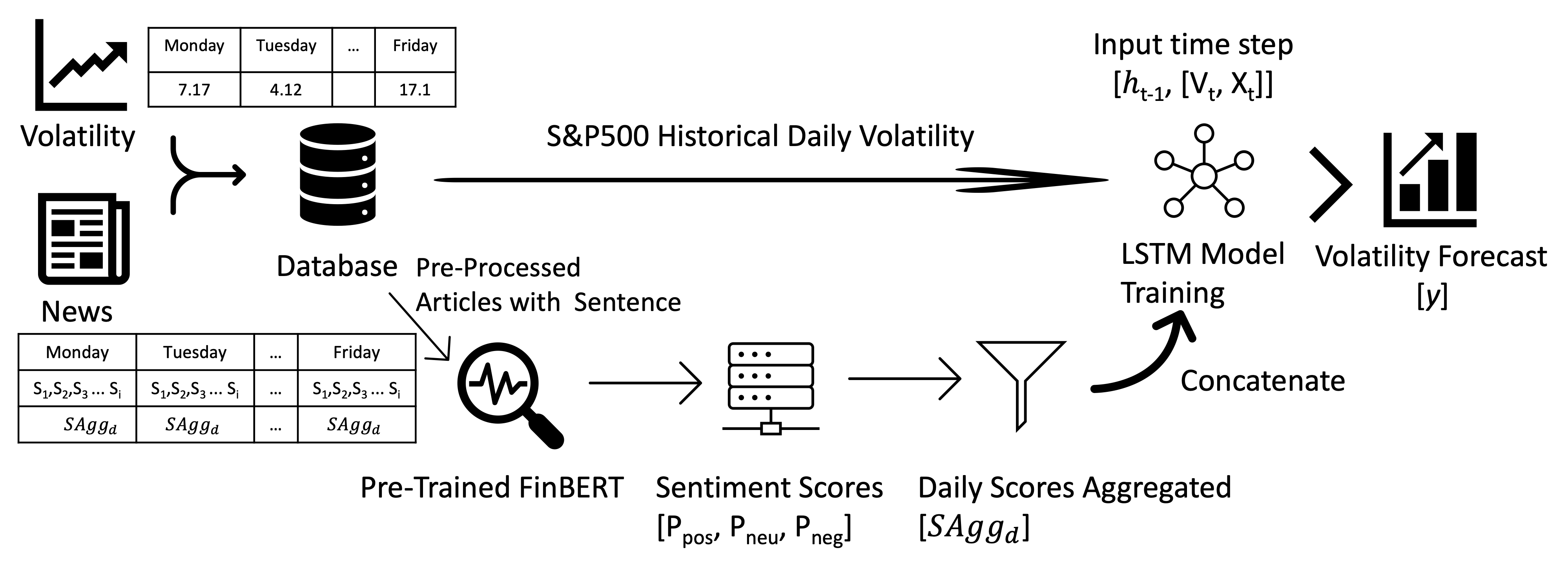}
\caption{The Proposed LSTM Experimental Flow Chart} \label{fig:fc}
\end{figure}

The experimental setup and procedure, depicted in Fig.~\ref{fig:fc}, commences with the collection of news articles and historical stock data. Sentiment is first annotated at the sentence level within each news article. This process involves utilizing FinBERT pretrained model to assign a sentiment score to each sentence. These scores are then aggregated to compute an overall sentiment score daily.

With market volatility and aggregated sentiment scores prepared, we then put them into our proposed LSTM model. The model accepts input through an initial layer configured to handle sequences of data, with the input shape specified as timesteps and dim, and the timesteps is set to 1 to process data in single time-step increments, and dim is from 1 expanded to 2 to accommodate the inclusion of both volatility and sentiment scores as features. Sequentially, three LSTM layers are employed, each comprising 100 neurons. We've applied 2 dropout layers with a rate of 0.5 to prevent overfitting. The model is compiled using the Adam optimizer and mean squared error (MSE) as loss function. Training is conducted with 100 epochs, with a batch size of 32. After model trained, the test dataset proceeds to generate predictions for volatility and compared with baseline model without sentiment score.


\begin{figure}
    \centering
    \begin{minipage}[t]{0.3\textwidth}
        \centering
        \includegraphics[width=\linewidth]{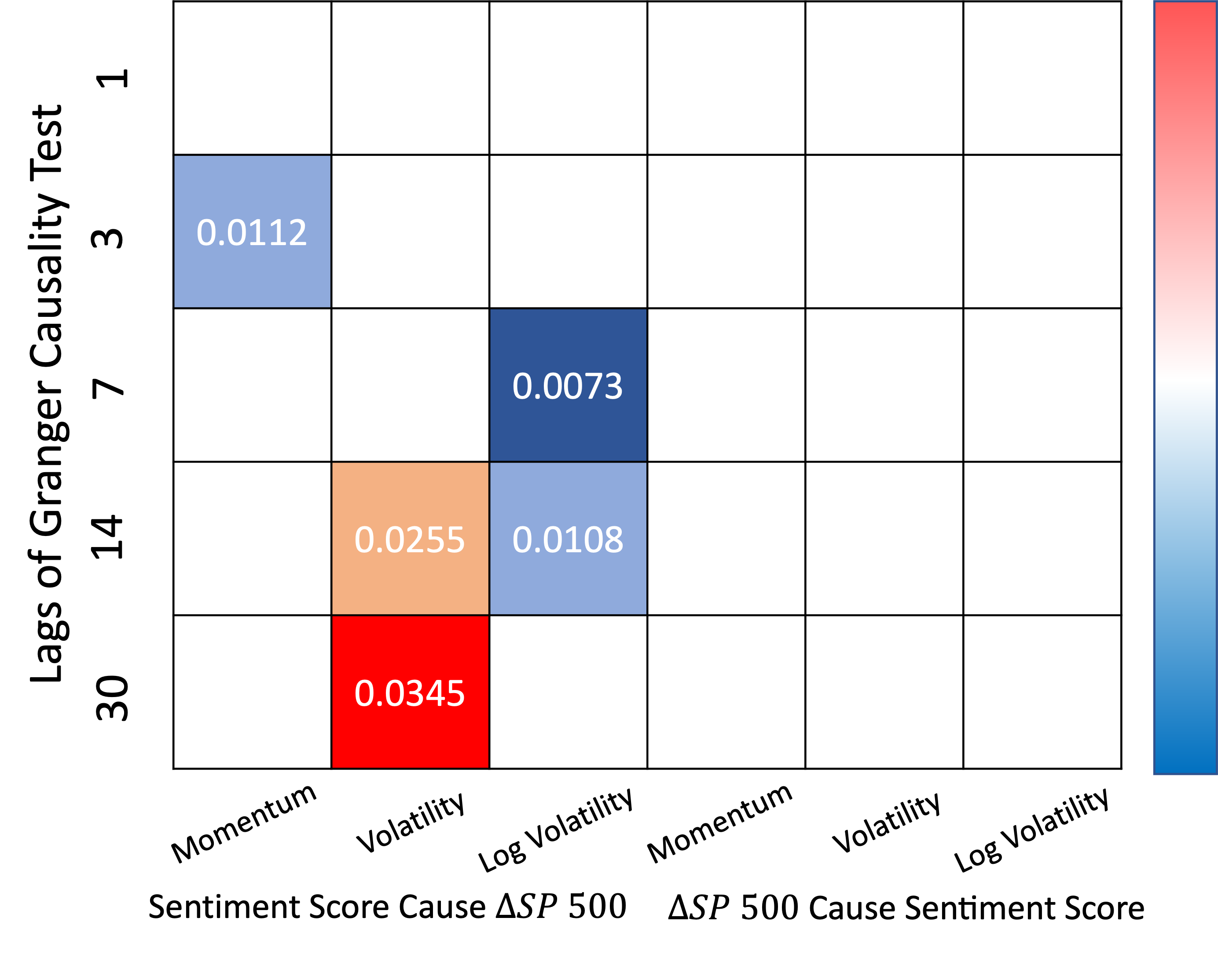}
        \caption{Granger Causality Test Result. Filtered \(p-values\) < 0.05 on lags 1, 3, 7, 14, 30.}
        \label{fig:granger}
    \end{minipage}\hfill
    \begin{minipage}[t]{0.66\textwidth}
        \centering
        \includegraphics[width=\linewidth]{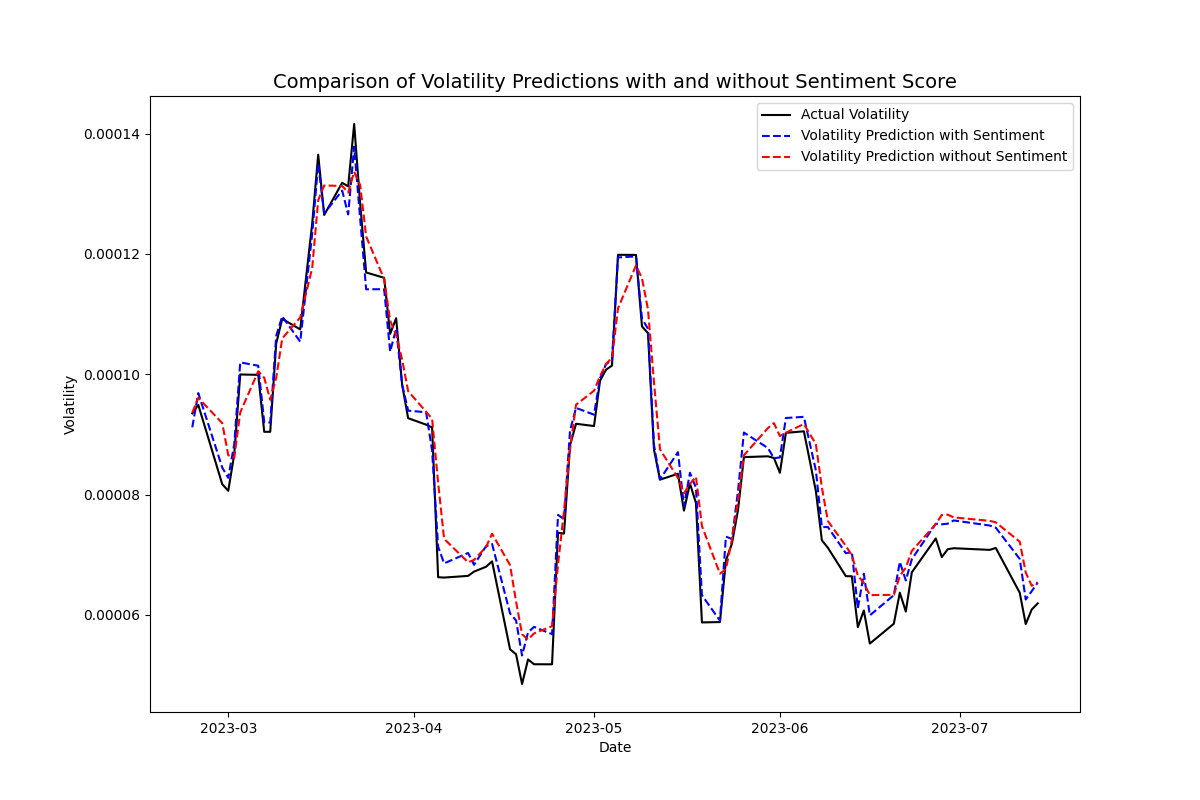}
        \caption{Volatility Prediction Comparison by LSTM Model. Red dash line is the prediction w/o sentiment score, while the blue dash line with it which is more closer with actual volatility (black line).}
        \label{fig:vol_pre}
    \end{minipage}
\end{figure}

The volatility prediction result over a period from March to July 2023 by LSTM model show in Fig.~\ref{fig:vol_pre}. The prediction with sentiment score is better than that one without and blue one is more closer approximation to the ground true value. Lag 14 exhibits better performance with lower error metrics (MAE, MSE, RMSE) and the highest R\(^2\) value 94.84\% among all lags. This implies that the model is most effective and accurate in capturing the relationship at this lag. The prediction result shows the evidence that the causal-validated sentiment score, as reflected in financial news, can help predict stock market volatility. 
\subsection{Calibration: Embedding Enhanced DAN 3 Model}
We used the 20 Newsgroups dataset  \cite{joachims1997probabilistic}, Financial PhraseBank \cite{malo2014good}, AG News \cite{zhang2015character} and FinSen(Ours) to train global-pooling CNN network \cite{kim2014convolutional} and DAN 3 \cite{guo2017calibration}. The datasets are preprocessed to load texts and their corresponding labels, tokenize the texts using a basic English tokenizer, and build a vocabulary from the tokenized texts. In the implementation, the GloVe (Global Vectors for Word Representation) \cite{pennington2014glove} embeddings are utilized within DAN 3 text classification model.  This GloVe embeddings used are from the 6B tokens dataset, which contains embeddings for 400k words, each represented in a 100-dimensional space. To accommodate varying lengths of text data, the texts are vectorized and padded to a uniform length, ensuring that each input to the model has the same size. This vectorization process transforms the tokenized texts into sequences of indices, with each index representing a word in the vocabulary. Padding is applied to shorter texts to reach the predefined maximum sequence length. Our models are trained on these datasets using cross entropy (CE), AdaFocal \cite{ghosh2022adafocal}, DualFocal \cite{tao2023dual} and Focal Calibration Loss (Ours) with 20 epochs and measured by Expected Calibration Error (ECE) \cite{guo2017calibration} and reliability diagram \cite{murphy1977reliability}. 

\begin{figure}[ht]
\centering
\includegraphics[width=0.85\textwidth]{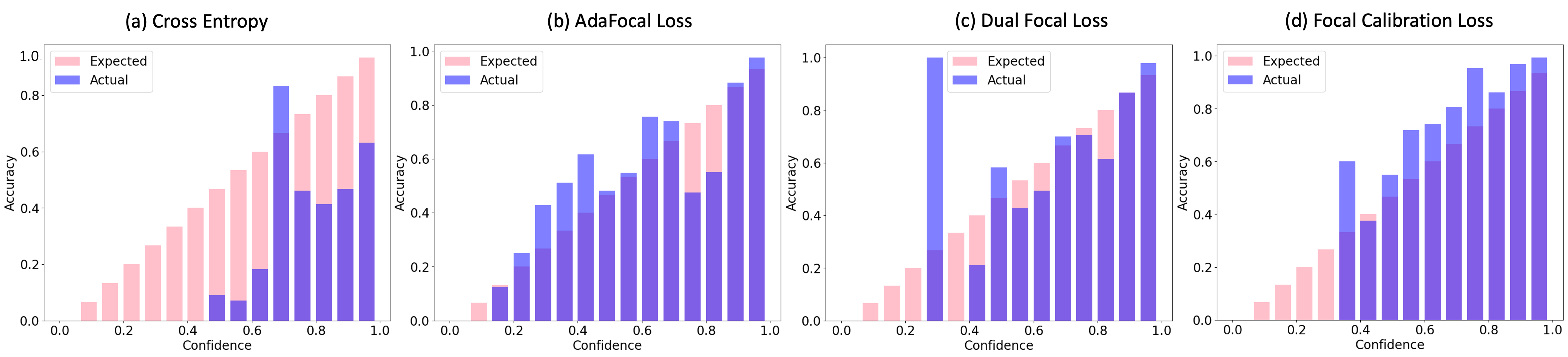}
\caption{FinSen Reliability Diagram of Different Methods w/o Temperature Scaling, \(\gamma\) setting for AdaFocal range from 1 to 20; \(\gamma\) for DFL is 2; \(\gamma\) for FCL is 2 and \(\lambda\) is 0.1.} 
\label{fig:dfl_finsen}
\end{figure}

Our FinSen dataset trained on DAN 3 model with Focal Calibration Loss shows an improvement in reliability diagram, which is competitive with other datasets in Fig.~\ref{fig:dfl_finsen}. With the DAN 3 model, FinSen achieves one of the lowest ECEs across all datasets and models, indicating that predictions are well-calibrated. In terms of ECE, the improvement for the FinSen dataset is notable, especially considering the significant reduction in classification error, which is only 0.13\% in global pooling CNN.

\begin{table}
\centering
\caption{Error metrics (\%) on test set}
\label{tab:ce_ece_result}
\resizebox{0.8\textwidth}{!}{%
\begin{tabular}{|c|c|c|c|c|c|c|}
\hline
\textbf{Datasets} & \textbf{Model} & \textbf{Error Type} & \textbf{CE} & \textbf{AFL} & \textbf{DFL} & \textbf{FCL(Ours)} \\
\hline

\multirow{3}{*}{\textbf{FinSen}} & \multirow{3}{*}{Global Pooling CNN} & Classification & 4.73 & 2.80 & 0.95& \textbf{0.13} \\
\cline{3-7}
& & ECE & 6.46 & 6.95 & 5.55 & \textbf{0.76}\\
\cline{3-7}
& & MCE & 47.70 & 53.90 & 76.21 & \textbf{42.51}\\
\hline

\multirow{3}{*}{\textbf{FinSen}} & \multirow{3}{*}{DAN 3} & Classification & 39.00 & 33.92 & 15.32 & \textbf{4.43} \\
\cline{3-7}
& & ECE & 36.97 & 13.94 & 4.77 & \textbf{3.34} \\
\cline{3-7}
& & MCE & 49.58 & \textbf{30.06} & 66.67 & 33.29 \\
\hline
\end{tabular}
}
\end{table}
\section{Conclusion}
By proving a causal relationship between sentiment scores and market volatility and integrating these causal-validated feature into LSTM models, our model achieves more nuanced and reliable market forecasts. Our FinSen dataset demonstrates relative better calibration on text classification task among others. The dramatic improvement in ECE to 3.34\% with the DAN 3 model on our Focal Calibration Loss suggests that not only has the model achieved a high level of accuracy, but its probability predictions are also well-aligned with actual outcomes, making it highly reliable for decision-making processes.

\section*{Appendix}

\appendix 
\renewcommand{\thesection}{\arabic{section}}

\section{Momentum, Volatility and Log Volatility}
\label{appendix:formulas}

This appendix provides the detailed mathematical formulations used in our analysis to calculate the momentum and volatility of financial assets. \(\text{Momentum}_t = P_t - P_{t-n}\) represents the calculation of momentum at time \(t\), where \(P_t\) is the price of the asset at time \(t\), and \(P_{t-n}\) is the price of the asset \(n\) periods before \(t\). Equation \(\text{Volatility} = \sqrt{\frac{1}{N-1} \sum_{i=1}^{N} (R_i - \bar{R})^2}\) defines the volatility of an asset, calculated as the standard deviation of its returns. \(R_i = \left(\frac{P_t}{P_{t-n}} - 1\right) \times 100\%\) represents the return of the asset at time \(i\), and \(\bar{R}\) is the average return over \(N\) periods. Equation \(\text{Log Volatility} = \ln(\text{Volatility})\) shows the natural logarithm of the volatility, which log volatility used in certain financial models.

\section{Evaluation Metrics \& Other Calibration Experiments}
\begin{description}
    \item[Mean Absolute Error (MAE)] $ = \frac{1}{n}\sum_{i=1}^{n}|y_i - \hat{y}_i|$
    \item[Mean Squared Error (MSE)] $ = \frac{1}{n}\sum_{i=1}^{n}(y_i - \hat{y}_i)^2$
    \item[Root Mean Squared Error (RMSE)] $ = \sqrt{\frac{1}{n}\sum_{i=1}^{n}(y_i - \hat{y}_i)^2}$
    \item[Mean Absolute Percentage Error (MAPE)] $ = \frac{1}{n}\sum_{i=1}^{n}\left|\frac{y_i - \hat{y}_i}{y_i}\right| \times 100\%$
    \item[Coefficient of Determination ($R^2$)] $ = 1 - \frac{\sum_{i=1}^{n}(y_i - \hat{y}_i)^2}{\sum_{i=1}^{n}(y_i - \bar{y})^2}$
    \item[Akaike Information Criterion (AIC)] $ = 2k - 2\ln(L)$
    \item[Bayesian Information Criterion (BIC)] $ = \ln(n)k - 2\ln(L)$
    \item[Expected Calibration Error (ECE)] $ = \sum_{m=1}^{M} \frac{|B_m|}{n} \left| \text{acc}(B_m) - \text{conf}(B_m) \right| $ 
    \item[Maximum Calibration Error (MCE)] $ = \max_{m=1}^{M} \left| \text{acc}(B_m) - \text{conf}(B_m) \right|$

\end{description}

\begin{table}[H]
\centering
\caption{Error metrics (\%) on test set}
\label{tab:ce_ece_result_other3}
\resizebox{0.8\textwidth}{!}{%
\begin{tabular}{|c|c|c|c|c|c|c|}
\hline
\textbf{Datasets} & \textbf{Model} & \textbf{Error Type} & \textbf{CE} & \textbf{AFL} & \textbf{DFL} & \textbf{FCL(Ours)} \\
\hline
\multirow{3}{*}{\textbf{20 Newsgroups}} & \multirow{3}{*}{Global Pooling CNN} & Classification & 24.25 & 24.8 & 25.3 & \textbf{21.85}\\
\cline{3-7}
& & ECE & 6.82 & 6.24 & \textbf{5.55} & 11.13\\
\cline{3-7}
& & MCE & \textbf{17.28} & 19.65 & 20.29 & 30.07\\
\hline
\multirow{3}{*}{\textbf{20 Newsgroups}} & \multirow{3}{*}{DAN 3} & Classification & 83.00 & 84.25 & 80.00 & \textbf{77.65}\\
\cline{3-7}
& & ECE & 31.69 & 53.90 & \textbf{28.14} & 29.97\\
\cline{3-7}
& & MCE & 50.99 & 76.27 & \textbf{42.14} & 52.90\\
\hline
\multirow{3}{*}{\textbf{AG News}} & \multirow{3}{*}{Global Pooling CNN} & Classification & 73.60 & 72.93 &72.31 & \textbf{71.75}\\
\cline{3-7}
& & ECE & 45.57 & 38.81 & 41.16 & \textbf{29.04}\\
\cline{3-7}
& & MCE & 67.92 & 67.07 & 73.70 & \textbf{61.22}\\
\hline
\multirow{3}{*}{\textbf{AG News}} & \multirow{3}{*}{DAN 3} & Classification & 71.68 & \textbf{71.13} & 73.82 & 74.71\\
\cline{3-7}
& & ECE & 51.32 & \textbf{4.25} & 35.15 & 37.85\\
\cline{3-7}
& & MCE & 66.17 & 63.02 & \textbf{45.08} & 45.67\\
\hline
\multirow{3}{*}{\textbf{PhraseBank}} & \multirow{3}{*}{Global Pooling CNN} & Classification & 10.60 & 18.98 & \textbf{13.25} & 15.67\\
\cline{3-7}
& & ECE & \textbf{3.53} & 21.81 & 8.22 &12.08\\
\cline{3-7}
& & MCE & 21.70 & 89.22 & \textbf{18.63 } & 60.53\\
\hline

\multirow{3}{*}{\textbf{PhraseBank}} & \multirow{3}{*}{DAN 3} & Classification & 22.96 & 24.06 & 18.76 & \textbf{12.80}\\
\cline{3-7}
& & ECE & \textbf{3.94} & 4.15 & 11.31 & 4.74\\
\cline{3-7}
& & MCE & 61.32 & \textbf{19.30} & 32.43 & 56.49\\
\hline
\end{tabular}
}
\end{table}
%
%
%
\bibliographystyle{splncs04}
\bibliography{mybibliography}

%




\end{document}